\documentclass[a4paper]{article}
\pdfoutput=1
\usepackage{INTERSPEECH2021}
\usepackage{latexsym}
\usepackage{amsmath}
\usepackage{url}
\usepackage{amssymb}
\usepackage{amsfonts}
\usepackage{graphicx}
\usepackage{tabularx}
\usepackage{multirow}
\usepackage{arydshln}
\usepackage{mathtools,nccmath}
\usepackage[T5]{fontenc}
\usepackage{enumitem}
\usepackage{todonotes}
\usepackage[most]{tcolorbox}
\usepackage[hidelinks]{hyperref}
\definecolor{IM}{rgb}{0.0, 0.0, 1.0}
\definecolor{RM}{rgb}{1.0, 0.0, 0.0}

\newcommand{\annotateSL}[1]{
\begin{tcolorbox}[colback=IM!5!white,colframe=IM!75!black,hbox,title=airport\_name,on line,before upper={\rule[-3pt]{0pt}{5pt}},boxrule=1pt,
boxsep=0pt,left=2pt,right=2pt,top=2pt,bottom=.5pt]
  #1
\end{tcolorbox}
}
\newcommand{\annotateDF}[1]{
\begin{tcolorbox}[colback=RM!5!white,colframe=RM!75!black,hbox,title=DF,on line,before upper={\rule[-3pt]{0pt}{5pt}},boxrule=1pt,
boxsep=0pt,left=2pt,right=2pt,top=2pt,bottom=.5pt]
   #1
 \end{tcolorbox}
}

\title{From Disfluency Detection to Intent Detection and Slot Filling}

\name{Mai Hoang Dao$^1$, Thinh Hung Truong$^{2}$, Dat Quoc Nguyen$^1$}
\address{$^1$VinAI Research, Vietnam; $^2$The University of Melbourne, Australia}
\email{\{v.maidh3, v.datnq9\}@vinai.io; hungthinht@student.unimelb.edu.au}

\begin{document}
\maketitle
\begin{abstract} 
We present the first empirical study investigating the influence of disfluency detection on downstream tasks of intent detection and slot filling. We perform this study for Vietnamese---a low-resource language that has no previous study as well as no public dataset available for disfluency detection. First, we extend the fluent Vietnamese intent detection and slot filling dataset PhoATIS by manually adding contextual disfluencies and annotating them. Then, we conduct experiments using strong baselines for disfluency detection and joint intent detection and slot filling, which are based on pre-trained language models. We find that: (i) disfluencies produce negative effects on the performances of the downstream intent detection and slot filling tasks, and (ii) in the disfluency context, the pre-trained multilingual language model XLM-R helps produce better intent detection and slot filling performances than the pre-trained monolingual language model PhoBERT, and this is opposite to what generally found in the fluency context. 
\end{abstract}


\section{Introduction}
% introduction to disfluency
% important to downstream task
% limit of previous work

In natural conversations, humans sometimes inevitably produce interruptions in their speech, which is formally referred to as \textit{disfluency} \cite{godfrey1993switchboard, shriberg1994preliminaries}. Its characteristic that breaks an utterance's semantic and syntax structures might make negative effects on the performances of downstream spoken language understanding (SLU) tasks as SLU models are primarily trained on curated and cleaned input without disfluencies. Thus, disfluency detection that detects
(and then removes) disfluencies to produce fluent versions of disfluent inputs is crucial in real-world applications. Most previous works study the disfluency detection task isolatedly \cite{hough15_interspeech, zayats16_interspeech,jamshid-lou-etal-2018-disfluency,wang-etal-2017-transition,bach2019noisy} and evaluate the task using gold disfluency annotations \cite{godfrey1993switchboard}, while investigation of this task's influence on downstream tasks is relatively limited. In particular, downstream tasks explored with disfluency contexts include punctuation restoration \cite{wang-etal-2014-combining,Lin2020}, machine translation \cite{salesky-etal-2019-fluent,wangetal}, syntactic parsing \cite{yoshikawa-etal-2016-joint, honnibal-johnson-2014-joint, rasooli-tetreault-2013-joint,jamshid-lou-johnson-2020-improving} and  question answering \cite{gupta-etal-2021-disfl}. Given the increasing popularity of task-oriented dialogue systems, it is naturally reasonable to ask a question on how disfluencies affect two important downstream SLU tasks of intent detection and slot filling.

To the best of our knowledge, no study has investigated the effect of disfluencies on the intent detection and slot filling tasks. The main reason is that there is no available dataset containing linguistic annotations over both disfluencies, intents, and the slots of the intents; and creating such a dataset is required to answer the question above. Inspired by Gupta et al. \cite{gupta-etal-2021-disfl} who present a disfluent derivative of the question answering dataset  SQUAD \cite{rajpurkar-etal-2016-squad}, a possible strategy to create a disfluent intent detection and slot filling dataset is to manually add contextual disfluencies into an existing fluent intent detection and slot filling dataset. This process could be performed for English with many publicly available intent detection and slot filling datasets \cite{price-1990-evaluation,coucke2018snips}. However, from a societal, linguistic, machine learning, cognitive, cultural, and normative perspective \cite{donlpotherlanguages}, it is also worth studying the proposed question for languages other than English, e.g. Vietnamese. %In particular, it is interesting to study whether the difference in linguistic characteristics might add difficulties to answering the question, e.g. investigating the influence of Vietnamese word segmentation \cite{vnwordseg}. 
Despite being the 17th most spoken language in the world \cite{Ethnologue} with about 100M speakers, Vietnamese is a low-resource language w.r.t. SLU tasks, e.g. having no previous study as well as no public dataset available for disfluency detection. 

In this paper, we present the first study that investigates the influence of disfluency detection on the downstream intent detection and slot filling tasks. We perform this study for Vietnamese---a low-resource language in these SLU research topics. First, we create a dataset with disfluency annotations by manually adding contextual disfluencies as distractors into the fluent dataset PhoATIS \cite{dao21_interspeech} which is the only current dataset publicly available for Vietnamese intent detection and slot filling. Then, we formulate our empirical approach as a ``Cascaded'' one combining a disfluency detection model and a joint intent detection and slot filling model. We conduct experiments using strong baseline models that are based on pre-trained language models {XLM-R} \cite{conneau2019unsupervised} and {PhoBERT} \cite{nguyen2020phobert}. Experimental results show that: (i) disfluencies negatively affect the performances of the downstream intent detection and slot filling tasks, and (ii) in the disfluency context, the pre-trained multilingual language model XLM-R is more effective for the intent detection and slot filling tasks than the pre-trained monolingual language model PhoBERT, and this is completely opposite to what is generally found in the fluency context with other Vietnamese NLP tasks \cite{nguyen2020phobert,vitext2sql,PhoNER_COVID19}. 

We publicly release our dataset with disfluency annotations to facilitate future Vietnamese SLU research and applications. Our dataset is available at \url{https://github.com/VinAIResearch/PhoATIS_Disfluency}.

\begin{table*}[!t]
    \centering
    \caption{A fluent utterance example and its disfluent variant with an intent label of ``ground\_service''.}
    \def\arraystretch{1.25}
    \begin{tabular}{p{0.85\linewidth}}
    %\hline
    %\textbf{Intent}: ground\_service \\
    %\hline
    \hline
    \textbf{Fluent utterance}:
    các phương tiện giao thông đường bộ có hoạt động ở \annotateSL{sân bay indianapolis} không \\
    \textit{English translation}: 
    is there ground transportation available at the \annotateSL{airport of indianapolis} \\
    \hline
    \textbf{Disfluent variant}:
    \annotateDF{giúp tôi tìm hạng vé à mà thôi} các phương tiện giao thông đường bộ có hoạt động \annotateDF{ở thành phố ờ không} ở {sân bay \annotateDF{ờ indapolis ý tôi là} indianapolis} không\\
    \textit{English translation}: 
    \annotateDF{please the find ticket class no actually} is there ground transportation available \annotateDF{in the city uh no} at the airport of \annotateDF{uh indapolis no i mean}  indianapolis  \\
    \hline
    \textbf{Disfluency terms break a slot's span}: \\
    \annotateDF{giúp tôi tìm hạng vé à mà thôi} các phương tiện giao thông đường bộ có hoạt động \annotateDF{ở thành phố ờ không} ở \annotateSL{sân bay \annotateDF{ờ indapolis ý tôi là} indianapolis} không\\
    \annotateDF{please find the ticket class no actually} is there ground transportation available \annotateDF{in the city uh no} at the \annotateSL{airport of \annotateDF{uh indapolis no i mean}  indianapolis} \\
    \hline
    \end{tabular}
%}
    \label{tab:example}
\end{table*}

\section{Our dataset} \label{sec:dataset}

To create our dataset, our approach is to manually add contextual disfluencies as distractors into the intent detection and slot filling dataset PhoATIS which consists of 5871 fluent utterances. 

Note that most work on automatic disfluency detection is aimed at removing disfluent Reparandum and Interregnum words to obtain fluent versions of input utterances for further processing as stated in  \cite{zayats16_interspeech,ostendorf13_interspeech}. Here, the Reparandum represents word(s) that the speaker intends to delete, while the (optional) Interregnum represents filled pauses, discourse marker, and the like \cite{shriberg1994preliminaries}. For example, in the utterance ``tôi cần một chuyến bay đến \textcolor{purple}{hà nội} \textcolor{cyan}{à không} {hải phòng} vào thứ tư tuần này'' ( i need a flight to ha noi uh no hai phong on this wednesday): ``\textcolor{purple}{hà nội}'' (ha noi) and ``\textcolor{cyan}{à không}'' (uh no)  can be labeled with Reparandum and Interragnum types, respectively. Recall that we aim to investigate the influence that disfluencies cause on models' ability to predict intent and slot labels. Thus, we do not separate these Reparandum and Interragnum types and merge them into a single type of ``Disfluency'' (denoted by \textbf{DF}). Revisiting the previous example, the whole disfluent phrase ``\textcolor{red}{hà nội à không}'' (ha noi uh no) is now labeled with DF. This strategy helps the models focus on the main tasks of intent detection and slot filling while still capable of detecting disfluencies.

We split the PhoATIS's training set into 5 equal and non-overlapping subsets and preserve its validation and test sets. We thus have 7 subsets that are used for crafting disfluencies. We employ 7 annotators who are undergraduate students strong in linguistics to generate a disfluent version of each original fluent utterance by adding disfluent words (here, each annotator annotates a subset, paid 0.08 USD per sentence). The disfluent version should satisfy the following requirements: (i)  semantically equivalent to the original one; (ii)  natural in terms of human usage, grammatical errors, and meaningful distractors (i.e. the added disfluent words exist in real-world circumstances); (iii) containing disfluent words that are corrected by following intent or slot value keywords in the original utterance; and (iv) containing both disfluent Reparandum- and Interragnum-type words where possible.

\begin{table}[!t]
\centering
\caption{Dataset statistics. (1): The number of utterances. (2): The number of disfluency (DF) annotations. (3): The number of slot annotations projected from PhoATIS. (4): The number of slots where disfluent words break a slot’s span. (5), (6), and (7) denote the average lengths (i.e. numbers of syllable tokens) of an utterance, a DF annotation and a slot, respectively.
}

\begin{tabular}{l | l | l | l | l}
\hline
\textbf{Statistics} & \textbf{Train} & \textbf{Valid.} & \textbf{Test} & \textbf{All}\\
\hline
(1) \# Utterances & 4478 & 500 & 893 & 5871 \\
\hline
(2) \# DF & 5178 & 841 & 1123 &  7142\\
\hline
(3) \# Slots & 14859 & 1713 & 2842 & 19414 \\
\hline
(4) \# Slots w/ DF & 225 & 18 & 30 & 273 \\
\hline
(5) Avg. Utt. length & 22.1 & 24.1 & 22.2 & 22.3 \\
\hline
(6) Avg. DF length & 5.53 & 5.14 & 6.14 &  5.58 \\
\hline
(7) Avg. slot length & 2.13 & 1.96 & 2.03 & 2.1 \\
\hline
\end{tabular}
%}
\label{tab:statistics}
\end{table}

The annotators are shown example disfluencies as illustrated in Table \ref{tab:example}. They are also required to make sure that the exact original utterance can be obtained when removing all the added words in the disfluent version. Once the adding process is completed, the first two authors manually revisit each utterance to ensure that all the requirements are met, discuss ambiguous cases and make further revisions if needed. This process results in a dataset of 5871 disfluent utterances, where each phrase spanning over continually added words is labeled with DF. When projecting slot annotations from the fluent PhoATIS dataset into our disfluent dataset, we find 273 cases where disfluent words break a slot’s span, as illustrated in Table \ref{tab:example}. Table \ref{tab:statistics} reports other statistics of our dataset.

\textbf{Note that} when written in Vietnamese texts, the white space is used as the delimiter between words and also as the delimiter between syllables that constitute a word. Thus, the annotation process is performed at the syllable level for convenience (e.g. the example in Table \ref{tab:example}). To obtain a word-level variant of the dataset, we employ RDRSegmenter \cite{NguyenNVDJ2018} from the VnCoreNLP toolkit \cite{vncorenlp} to perform automatic Vietnamese word segmentation.  For example, a 6-syllable written text ``sân bay quốc tế Nội Bài'' (Noi Bai international airport) is word-segmented into a 3-word text ``sân\_bay\textsubscript{airport} quốc\_tế\textsubscript{international} Nội\_Bài \textsubscript{Noi\_Bai}''.

%\begin{figure}[!]
%    \centering
%    \includegraphics[width=0.475\textwidth]{JointModel.png}
%    \caption{Our model architecture for jointly learning three tasks of disfluency detection, intent detection and slot filling.}% Without the disfluency detection component, our model reduces to JointBERT+CRF.}
%    \label{fig:jointmodel}
%\end{figure}

%\begin{figure}[!t]
%   \centering
%    \includegraphics[width=0.475\textwidth]{DisfluencyDetection_diagram.pdf}
%    \caption{Illustration of the disfluency detection architecture.}% Without the disfluency detection component, our model reduces to JointBERT+CRF.}
%    \label{fig:disfluency}
%\end{figure}

%\begin{figure}[!t]
%   \centering
%    \includegraphics[width=0.475\textwidth]{JointBERT_diagram.pdf}
%    \caption{Illustration of the JointBERT+CRF architecture.}% Without the disfluency detection component, our model reduces to JointBERT+CRF.}
%    \label{fig:jointbert}
%\end{figure}

\section{Empirical approach} \label{sec:approach}

Due to the nature of our Vietnamese dataset, where disfluent terms might break a slot's span as illustrated in Table \ref{tab:example}, we study the impact of disfluency detection on downstream intent detection and slot filling tasks using a ``{Cascaded}'' approach.

\subsection{Modeling}

Our ``Cascaded'' approach combines two separate models: (i) disfluency detection and (ii) joint intent detection and slot filling (here, it is worth noting that jointly learning these two tasks of intent detection and slot filling helps improve performance results compared to the single-task training \cite{louvan2020recent,zhang-etal-2019-joint,Weld2021ASO}). % We thus treat the disfluency detection component as a pre-processing module to help improve the readability of the input. 
In particular, given an input utterance, we first use the disfluency detection model to automatically identify disfluent terms and then remove these identified terms to generate a ``fluent'' variant---i.e. a version with automatic disfluency removal---of the input. We feed the ``fluent'' variant into the joint intent detection and slot filling model to predict intent and slot types. 

Previous studies show that the sequence labeling (i.e. token classification) strategy that fine-tunes pre-trained language models (LMs) produces state-of-the-art disfluency detection performances for English \cite{bach2019noisy,rocholl21_interspeech}. In addition, fine-tuning pre-trained LMs also help produce state-of-the-art performances for other Vietnamese sequence labeling tasks  \cite{nguyen2020phobert,PhoNER_COVID19}. 
Thus we formulate the Vietnamese disfluency detection task as a sequence labeling problem with the frequently used tagging scheme BIO (here, the label set for disfluency detection consists of  B-DF, I-DF, and O only). The disfluency detection model %, as illustrated in Figure \ref{fig:disfluency}, 
employs a pre-trained LM-based encoder to generate contextualized latent feature embeddings for the input tokens. Each latent feature embedding is then linearly transformed before being fed into a linear-chain CRF layer \cite{Lafferty:2001} for disfluency label prediction.

The joint intent detection and slot filling model we employ is JointBERT+CRF \cite{jointbert} which also formulates slot filling as a sequence labeling problem and obtains state-of-the-art performances for Vietnamese intent detection and slot filling \cite{dao21_interspeech}. JointBERT+CRF %, as illustrated in Figure \ref{fig:jointbert}, 
inserts a special classification token of ``[CLS]'' at the front of its input token sequence. Then it also employs a pre-trained LM-based encoder to generate contextualized latent feature embeddings. %Following a common manner when fine-tuning pre-trained LMs for a sequence classification task \cite{devlin-etal-2019-bert}, 
JointBERT+CRF appends a linear prediction layer---i.e. a single-layer feed-forward network followed by a $\mathsf{softmax}$ predictor---on top of the contextualized embedding of the classification token  ``[CLS]'' for intent detection. The remaining contextualized embeddings are linearly transformed before being fed into a linear-chain CRF layer for slot type prediction.

\subsection{Implementation details} \label{subsec:exsetup}

We train the disfluency detection model using the training set of disfluent utterances with disfluency annotations only (see the disfluent variant example in Table \ref{tab:example}), while we train JointBERT+CRF for joint intent detection and slot filling using the gold fluent PhoATIS training set, i.e. with only intent and slot annotations  (see the fluent utterance example in Table \ref{tab:example}).

Recall that input utterances can be represented at either the syllable or word level. For the syllable-level input, our pre-trained LM-based encoder is {XLM-R} \cite{conneau2019unsupervised}, while it is {PhoBERT} \cite{nguyen2020phobert} for the word-level input. Here, XLM-R and PhoBERT are multilingual and Vietnamese monolingual variants of the language model RoBERTa \cite{RoBERTa}. XLM-R is pre-trained on a 2.5TB multilingual dataset that contains 137GB of syllable-level Vietnamese texts, while PhoBERT is pre-trained on a 20GB word-level Vietnamese corpus.

We implement models using PyTorch \cite{paszke2019pytorch}, employing pre-trained XLM-R and PhoBERT available from \texttt{transformers} \cite{wolf-etal-2020-transformers}. 
For each model, we train for 50 epochs,  employ the AdamW optimizer \cite{loshchilov2018decoupled} and set the batch size to 32. We also perform grid search on the validation set to select the optimal Adam initial learning rate in \{1e-5, 2e-5, 3e-5, 4e-5, 5e-5\}.  
We calculate the F\textsubscript{1}-score (in \%) of the disfluency detection model after each training epoch on the disfluent validation set and select the model checkpoint that obtains the highest F\textsubscript{1}-score to apply to the disfluent test set. We then produce versions with automatic disfluency removal of the disfluent validation and test sets for downstream intent detection and slot filling evaluations. For JointBERT+CRF, compared against the gold PhoATIS validation and test sets, we calculate the average score of the intent accuracy for intent detection and the F\textsubscript{1}-score (in \%) for slot filling after each training epoch on the automatic-disfluency-removal validation version, and we select the model checkpoint that obtains the highest average score to apply to the automatic-disfluency-removal test version. All our reported results are the average over 5 runs with 5 different random seeds.

\begin{table}[!t]
\centering
\caption{Test set results. ``Dis.~F\textsubscript{1}'', ``Int. Acc.'', ``Slot~F\textsubscript{1}''  and ``Sen.  Acc.'' denote the F\textsubscript{1} score (in \%) for disfluency detection, the intent detection accuracy, the F1  score (in \%) for slot filling and the sentence-level accuracy, respectively. ``Gold'' denotes the use of the gold PhoATIS test set for intent detection and slot filling evaluation, i.e. equivalent to a perfect disfluency detection (F\textsubscript{1} at 100\%). ``Predicted'' denotes the evaluation that follows our empirical approach with predicted disfluencies.}

\def\arraystretch{1.1}
\setlength{\tabcolsep}{0.3em}
\resizebox{8cm}{!}{
\begin{tabular}{l|l|l|c|c|c|c}
\hline
\multicolumn{2}{c|}{\textbf{Mode}} & \textbf{Encoder}  &\textbf{Dis. F\textsubscript{1}}& \textbf{Int. Acc.}  & \textbf{Slot F\textsubscript{1} }  & \textbf{Sen. Acc.} \\
\hline
\multirow{2}{*}{\rotatebox[origin=c]{90}{\textbf{Syll.}}}
& Gold   & XLM-R & 100.0  & 97.42 & 94.62 & 85.39 \\
\cline{2-7}
& Predicted  & XLM-R  & 93.85 & 97.20 & 94.11  &	84.21 \\
\cline{2-7}
%& JointDF & XLM-R  &	94.07 &	86.76 &	97.36 &	55.66 \\
\hline
\multirow{2}{*}{\rotatebox[origin=c]{90}{\textbf{Word}}}
& Gold  & PhoBERT & 100.0  & 97.40 & 94.75 & 85.55 \\
\cline{2-7}
& Predicted  & PhoBERT  & 94.33 &	97.31 & 93.37 & 81.74 \\
\cline{2-7}
%& JointDF & PhoBERT & 94.30	& 85.92	& 97.31 & 52.30 \\
\hline
\end{tabular}
}
\label{tab:mainresults}
\end{table}

\section{Experimental results}

\subsection{Main results}

Table \ref{tab:mainresults} reports obtained results on the test set, including the F\textsubscript{1} score for disfluency detection as well as the intent accuracy for intent detection, the F\textsubscript{1} score for slot filling, and the sentence-level accuracy w.r.t. both intent detection and slot filling.  We categorize the results into two comparable settings based on the syllable- and word-level types of input utterances, associated with the encoders XLM-R and PhoBERT, respectively.

For disfluency detection, the model that employs the word-level input with PhoBERT encoder obtains a higher F\textsubscript{1} score than the one employing the syllable-level input with XLM-R encoder (i.e. 94.33\% vs. 93.85\%). This seems reasonable as syllables constitute words, resulting in disfluent phrases at the syllable level which are ``longer'' w.r.t. the average number of tokens, than those at the word level, e.g. a 4-syllable phrase ``\textcolor{red}{hà nội à không}'' vs. a 3-word phrase ``\textcolor{red}{hà\_nội à không}''. The model thus likely finds it more difficult to predict exact annotation boundaries of the ``longer'' disfluent phrases at the syllable-level.

When it comes to the effect of disfluency detection on the two downstream tasks, all downstream performance scores are decreased: 97.42\%  $\rightarrow$ 97.20\% and 97.40 $\rightarrow$ 97.31\%, which are accuracies for intent detection at the syllable and word levels, respectively; and 94.62\% $\rightarrow$ 94.11\% and 94.75\% $\rightarrow$ 93.37\%, which are F\textsubscript{1} scores for slot filling at the syllable and word levels, respectively. It can be explained by errors propagation from the disfluency detection phase which generates utterances with missing fluent tokens and left-over disfluent tokens. Note that the absolute decreases in intent detection are smaller than those in slot filling. This is not surprising because intent detection is a sequence classification task while slot filling is formulated as a token classification task, thus slot filling errors might not induce an intent detection error. Final sentence-level accuracies witness substantial drops of 1.18\% (85.39\% $\rightarrow$ 84.21\%) for the syllable level and 3.81\% (85.55\% $\rightarrow$ 81.74\%) for the word level, illustrating the strong negative impact of disfluency detection on intent detection and slot filling.

Table \ref{tab:mainresults} shows that in the disfluency context, in general, the word-level model employing PhoBERT w.r.t. automatic Vietnamese word segmentation produces lower intent detection and slot filling performances than its syllable-level counterpart employing XLM-R. This is opposite to the obtained results with gold fluent input utterances in Table \ref{tab:mainresults} as well as to what is generally found with other Vietnamese NLP tasks in the fluent context, where PhoBERT does better than XLM-R \cite{nguyen2020phobert,vitext2sql,PhoNER_COVID19}. 
One possible reason for this phenomenon is potential errors from the automatic word segmentation process, where disfluent syllables and fluent syllables that appear next to each other are segmented into one word-level token; and once being predicted, the whole disfluent words containing fluent syllables will be removed, which leads to a more information loss. 

% \todo{One possible reason is: word-level utterance likely sẽ bị mất nhiều thông tin hơn khi mình remove disfluencies (do remove whole word thay vì individual syllable token)}

\subsection{Error analysis}

On the validation set, we perform an analysis to investigate the source of errors using the syllable-level models with XLM-R. 

\subsubsection{Disfluency detection errors}
We find that 53 phrases are fluent but being predicted as disfluent phrases i.e. the false positive instances, and 44 real disfluent phrases are mis-detected or partly recognized i.e. the false negative instances. Disfluency detection errors are more likely to happen in relatively long sentences with multi-token disfluent phrases and ambiguous contexts.   
%TT: give an example of a long sentence with multiple disfluent terms

\begin{table}[!t]
    \centering
    \caption{Counts for error types on the validation set (average over the 5 different runs).}
    \def\arraystretch{1.1}
    \setlength{\tabcolsep}{0.3em}
    \resizebox{7.75cm}{!}{
    \begin{tabular}{p{8cm} | c}
        \hline
        Definition & \# Errors\\
        \hline
         Wrong Intent (\textbf{WI}): Predicted intent label is not the gold-annotated one. & 11 \\
        \hline
         Missing Slot (\textbf{MS}): A gold slot's span is not entirely or partly recognized. & 14\\
        \hline
         Spurious Slot (\textbf{SS}): A predicted slot matches a gold  O label. & 9\\
        \hline
         Wrong Boundary (\textbf{WB}): A predicted slot's span is partly overlapped with a gold slot's span, while the predicted slot's label type is the gold slot's. & 12 \\
        \hline
         Wrong Label (\textbf{WL}): The predicted slot has exact span boundary while having incorrect slot label. & 29 \\
        \hline
    \end{tabular}
    }
    \label{tab:error}
\end{table}

\subsubsection{Intent detection and slot filling errors}

We categorize the error cases into 5 different categories including WI, MS, SS, WB, and WL. The definition of each category and its number of error cases are listed in Table \ref{tab:error}. 

We find 11 error cases for the WI category and most of them are caused by the multi-intent labels (e.g. ``airfare\#flight'') since the model is likely to select the most clearly manifested or first appeared intent. For example, given the utterance ``cho tôi danh sách các chuyến tàu à đâu các chuyến bay vào ngày 27 tháng 12 từ đài bắc đến thái lan à không đến singapore và giá vé tương ứng'' (give me the list of cruises uh no flights on december 27 from Taipei to Thailand no actually to Singapore and their respective fare), it  is predicted with the intent of ``flight'' instead of the gold intent label ``airfare\#flight''. There are also WI cases induced by disfluency detection errors.  For instance, given an input ``bạn có thể cho tôi biết giá vé xe buýt à không ý tôi là các chuyến bay giữa huế và cà mau được không'' (could you please show me the fare of the buses uh no i mean the flights between Hue and Ca Mau), the disfluency detection model identifies fluent terms  ``giá vé'' (the fare) as disfluent terms (and thus also remove these terms in the automatic disfluency removal phase), leading to a wrong prediction of intent ``flight''.

There are 14 and 9 error cases counted for MS and SS categories, respectively. These two types of errors are generally caused by the ambiguities over slot types that rarely occur in the training set such as ``connect'' (36/14859  training slot values) or ``economy'' (34/14859). 
The WB category has 12 error cases that are mostly induced by multi-syllable slot values, especially the incorrectly removed fluent tokens, and the incorrectly preserved disfluent ones. For example, given an input utterance ``cho tôi biết các hãng vận tải à không hãng hàng không có các chuyến bay đến hoặc đi từ sân bay tân sơn nhất à đâu doncaster sheffield'' (show me the transportation uh no airlines for flights to or from the airport of Tan Son Nhat no actually Doncaster Sheffield), performing automatic disfluency removal produces an utterance variant of ``cho tôi biết các hãng hàng không có các chuyến bay đến hoặc đi từ sân bay tân sơn nhất sheffield'' (show me the airlines for flights to or from the Tan Son Nhat Sheffield airport), resulting in a WB error case with ``sân bay tân sơn nhất sheffield'' (Tan Son Nhat Sheffield airport) tagged with label ``fromloc.airport\_name''. Here, the correct slot value is ``sân bay doncaster sheffield'' (Doncaster Sheffield airport).

The last category WL---the most common error type---contains 29 error cases. These errors exist mostly because of the ambiguities between the ``departure'' part and the ``arrival'' part of an utterance since many utterance contexts are not explicitly specified. Consider the utterance ``hiển thị các chuyến bay không dừng một chiều từ new york đến hà nội vào một ngày thứ ba'' (show nonstop flights from New York to Ha Noi on a tuesday), it is confusing to determine whether ``thứ ba'' (tuesday) is the departure date or arrival date without a clearer context. In addition, some of those cases are also caused by disfluency detection errors. For example, given an input utterance ``tôi cần một chuyến bay từ phú quốc đến hà nội không ý là thành phố hồ chí minh và sau đó từ thành phố hồ chí minh đến singapore à không jakarta và từ jakarta đến hà nội'' (i need a flight from Phu Quoc to Ha Noi no actually Ho Chi Minh City and then Ho Chi Minh city to Singapore  uhm no Jakarta and from Jakarta to Ha Noi), the phrase ``đến singapore à không'' (to Singapore  uhm no) is detected as a disfluent one, resulting in a WL case where ``thành phố hồ chí minh jakarta'' (Ho Chi Minh Jakarta city) is labeled  as ``fromloc-city\_name''.

\section{Conclusion}

In this paper, we have presented the first empirical study investigating the influence of disfluency detection on two downstream SLU tasks of intent detection and slot filling. We manually add contextual disfluencies into the fluent Vietnamese intent detection and slot filling dataset PhoATIS. Our dataset is the first dataset with disfluency annotations for Vietnamese. We then conduct experiments under the ``Cascaded'' manner with strong pre-trained LM-based baseline models and perform detailed error analysis. Experimental results show that disfluencies cause substantial performance degradation in the intent detection and slot filling tasks, and the pre-trained monolingual LM  PhoBERT is less effective than the pre-trained multilingual LM XLM-R for intent detection and slot filling under the disfluency context. We hope that our dataset and findings will facilitate future Vietnamese SLU research and applications.

\bibliography{refs}
\bibliographystyle{IEEEtran}

\end{document}